\documentclass[conference]{IEEEtran}
\IEEEoverridecommandlockouts
\usepackage{cite}
\usepackage{amsmath,amssymb,amsfonts}
\usepackage{algorithmic}
\usepackage{graphicx}
\usepackage{textcomp}
\usepackage{xcolor}
\def\BibTeX{{\rm B\kern-.05em{\sc i\kern-.025em b}\kern-.08em
    T\kern-.1667em\lower.7ex\hbox{E}\kern-.125emX}}

\begin{document}

\title{WIP: Development of a Student-Centered Personalized Learning Framework to Advance Undergraduate Robotics Education\\

\thanks{This material is based upon work supported by the National Science Foundation under Grants \#DUE-2142360, \#IUSE-2142428, and \#IUSE-1730568. All study activities were supervised by ECU's IRB office.}

}



\author{\IEEEauthorblockN{Ponkoj Chandra Shill}
\IEEEauthorblockA{\textit{Computer Science \& Engineering} \\
\textit{University of Nevada, Reno}\\
Reno, Nevada, USA \\
ponkoj@nevada.und.edu}
\and
\IEEEauthorblockN{Rui Wu}
\IEEEauthorblockA{\textit{Computer Science} \\
\textit{East Carolina University}\\
E 5th Street, Greenville, NC 27858 \\
wur18@ecu.edu}
\and
\IEEEauthorblockN{Hossein Jamali}
\IEEEauthorblockA{\textit{Computer Science \& Engineering} \\
\textit{University of Nevada, Reno}\\
Reno, NV, USA \\
hossein.jamali@nevada.unr.edu}
\and
\IEEEauthorblockN{Bryan Hutchins}
\IEEEauthorblockA{\textit{SERVE Center} \\
\textit{University of North Carolina Greensboro}\\
bchutchi@uncg.edu}
\and
\IEEEauthorblockN{Sergiu Dascalu,  Frederick C. Harris, Jr., David Feil-Seifer}
\IEEEauthorblockA{\textit{Computer Science \& Engineering} \\\textit{University of Nevada, Reno}\\
\{dascalus, Fred.Harris, dave\}@cse.unr.edu}
}

\maketitle

\begin{abstract}

This paper presents a work-in-progress on a learning system that will provide robotics students with a personalized learning environment. This addresses both the scarcity of skilled robotics instructors, particularly in community colleges and the expensive demand for training equipment. The study of robotics at the college level represents a wide range of interests, experiences, and aims. This project works to provide students the flexibility to adapt their learning to their own goals and prior experience. We are developing a system to enable robotics instruction through a web-based interface that is compatible with less expensive hardware. Therefore, the free distribution of teaching materials will empower educators. This project has the potential to increase the number of robotics courses offered at both two- and four-year schools and universities. 
The course materials are being designed with small units and a hierarchical dependency tree in mind; students will be able to customize their course of study based on the robotics skills they have already mastered. 
We present an evaluation of a five module mini-course in robotics. Students indicated that they had a positive experience with the online content. They also scored the experience highly on relatedness, mastery, and autonomy perspectives, demonstrating strong motivation potential for this approach.

%

                                                          
\end{abstract}


\section{Introduction}

Robotics education can prepare students for career success. However, it can be very difficult to give students a robotics education at the community college or primarily undergraduate institution level if those institutions do not have any robotics-trained faculty. We are developing self-paced, online course materials, which could be deployed at a community college or a university. A personalized learning server could remotely offer robotics course content for campuses without local robotics experts. Each student can study their choice of critical robotics concepts, in the same classroom, assisted by a local instructor, and utilizing an online coding/lab environment.

In this proposed teaching method, the main jobs of an instructor are to make sure students reach educational milestones in every class, collect students' questions, and distribute back answers from the course module designer and course content advisory committee. We are inspired by self-determination theory \cite{deci2015selfdetermination}, which shows increasing students' autonomy can enhance their motivation and engagement. The overarching goal of this project is to make headway in resolving problems that threaten the expansion and accessibility of robotics education.  
We are studying solutions for accessibility issues such as the difficulty institutions have locating qualified professors to teach these cutting-edge robotics courses. 

In this work-in-progress paper, we describe the initial personalized learning environment development, course module design and a 5-module mini-course with an evaluation of the content with University students in a classroom setting. 

\section{Background}


The emergence of advanced robotics technologies such as autonomous vehicles, drones, and medical robots has created many job opportunities. Robotics technology can create new employment opportunities \cite{ford2015rise}. The development of robotics technology will lead to the creation of new jobs in industries such as manufacturing, software development, and even healthcare. Investments in robotics are likely to lead to net gains in employment, wages, and economic growth~\cite{bcg2015robotics}.
The use of industrial robots led to the creation of three to five million jobs globally in 2015, which increased the demand and created new jobs representing a 10-15\% increase in the number of jobs in industries that use robots \cite{ifr2017impact}.

\begin{figure}[ht]
\centering
\includegraphics[width=0.95\linewidth]{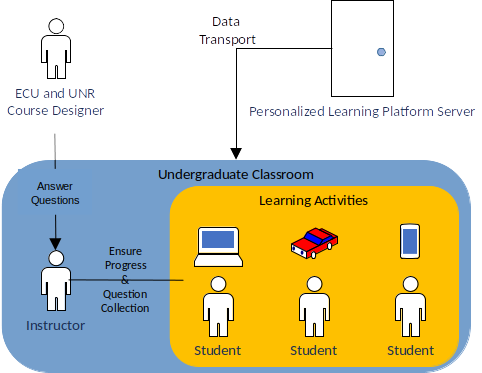}
\caption{Proposed student-centered personalized learning framework: students are at the center and an instructor is required to assist instead of leading the students. Instructors need to ensure students make progress in every class and collect questions from students. Students can work on different topics in the same classroom with different required hardware. This framework does not require students to have powerful devices.\label{fig:eduframework}}
\end{figure}

To effectively instruct on advanced robotics, community colleges, and universities require significant resources, including specialized hardware and proficient educators. Specialized hardware and proficient educators are necessary for teaching robotics because mobile robots present unique challenges that require a deep understanding of electronics, software development, and experimental methods~\cite{fasola2008robotics}. The environment, robot hardware, and software all play equally important roles in the behavior of a mobile robot, making it necessary to have specialized hardware and educators who can effectively teach students how to navigate these challenges~\cite{dias2007undergraduate}.  Additionally, the cost of robotics equipment can be prohibitive, and course content may quickly become obsolete due to the rapidly evolving nature of robotics \cite{miller2008robots}.

The gap between training ability and training need in undergraduate robotics education is a significant challenge, particularly in community schools and technologically under-served communities. Factors contributing to this issue include the insufficient availability of qualified robotics instructors, inadequate funding for equipment, and significant variations in the backgrounds and experiences of undergraduate students~\cite{dias2007undergraduate}. Training programs offered by major robotics companies may not be as beneficial as general robotics programs available at universities or community colleges. Financial barriers may exist for some students who cannot afford the cost of purchasing or renting robots~\cite{lauwers2009csbots}.  Establishing a robotics program may require substantial financial investments due to the need for specialized hardware and proficient educators \cite{miller2008robots}

Proficient educators are necessary for teaching robotics because robots are physically manifested computing devices that inherently show students how computing programs that they write can impact the real world~\cite{weinberg2005multidisciplinary}. However, the interdisciplinary nature of robotics can add a significant teaching challenge for instructors new to the field. Robots provide an opportunity for students to see how their programming skills can be applied in practical settings, which can be difficult to achieve with purely theoretical coursework. Additionally, specialized hardware is required because robots have unique physical characteristics and capabilities that must be taken into account when designing and programming them~\cite{lauwers2009csbots}.


This work in progress reduce the skills required to teach a robotics course so an instructor need not have multi-disciplinary engineering expertise~\cite{weinberg2005multidisciplinary}. By adapting the materials to be more accessible and providing support and resources for faculty members who may not have extensive experience in robotics education or research, costs can be reduced, and accessibility increased, allowing more institutions and students to participate in robotics education and training. This can lead to more diverse and skilled professionals entering the field. Reduced expenses help in reducing obstacles to participation in robotics classes by making it more affordable for students and institutions to offer and participate in such courses \cite{weinberg2005multidisciplinary}. Furthermore, reducing financial barriers enables students from different backgrounds to pursue their interests in robotics without worrying about the high costs associated with learning materials or equipment. 



To address these challenges, the authors propose the development of a customized learning framework that prioritizes individual students' needs in undergraduate robotics education. We aim to reduce the expenses required for developing a robotics curriculum and enable educators who lack expertise in robotics to instruct on the advanced subject matter.


\section{Approach}

Our proposed framework has three objectives:

\begin{enumerate}
\item 	Implement a student-centered personalized learning framework for hands-on robotics education;
\item 	Develop a mini-course in robotics utilizing this framework; and 
\item 	Conduct a study to evaluate the effectiveness of the proposed framework.
\end{enumerate}


The proposed teaching method will enhance undergraduate robotics education by offering students the freedom to choose their robotics learning path, while at the same time without requiring instructors to have robotics expertise. We want to develop an education framework (see Figure~\ref{fig:eduframework}) that does not require a robotics expert instructor in the classroom. 

\subsection{Student Learning Framework}

The proposed teaching method for the mini-course involves utilizing the ISPeL platform (see Figure~\ref{fig:eduframework}) developed by East Carolina Univerity (ECU). The course content, including videos and sample code wrapped in Jupyter-notebooks~\cite{kluyver2016jupyter}, can be accessed by students through the platform. Instructors are responsible for ensuring that students reach milestones in each class and collecting their questions. If the instructor is unable to answer a question, they can refer to a ``frequently asked questions and guidelines" document created by the course module designer and course content advisory committee. If the question remains unresolved, the instructor can consult with the course module designer and course content advisory committee via email. The main focus of the instructor is to facilitate student progress, collect and distribute answers, and manage devices correctly. The teaching method emphasizes the importance in-person classroom attendance to ensure progress monitoring, correct device usage, and collaborative knowledge sharing among students. 

To implement or start the mini-course, instructors can utilize the ISPeL platform hosted on an ECU server. They have the option to upload course content to the platform or customize and reuse topic components from another course. The instructor can organize the course content using a dependency graph, which is automatically generated when they order the topic components in a book chapter/sub-chapter style through simple mouse movements. This allows for visualizing the relationships and dependencies between different components. The dependency graph is based on the hierarchy defined by the instructor, ensuring a logical progression of learning.

\subsection{Topic Selection}

We developed the mini-course to establish this logical progression and interdependence among the selected topics. This implies that the concepts taught in earlier topics should serve as a basis for later ones. By establishing minimal dependencies, students can incrementally build upon their knowledge, resulting in a deeper understanding of the subject matter. If the course begins with an introduction to programming concepts, for instance, subsequent topics could concentrate on programming in the context of robotics, such as controlling robot movements or integrating sensors. Additionally, Given the short duration of the mini-course and the desire to enable students to choose topics based on their interests, it is essential to minimize topic dependencies. This ensures that students can enlist in individual courses without feeling overwhelmed or disadvantaged if they have not completed prerequisite courses. By reducing dependencies, students are able to select topics that correspond to their specific interests.

\begin{figure}[t]
\centering
\includegraphics[width=0.95\linewidth]{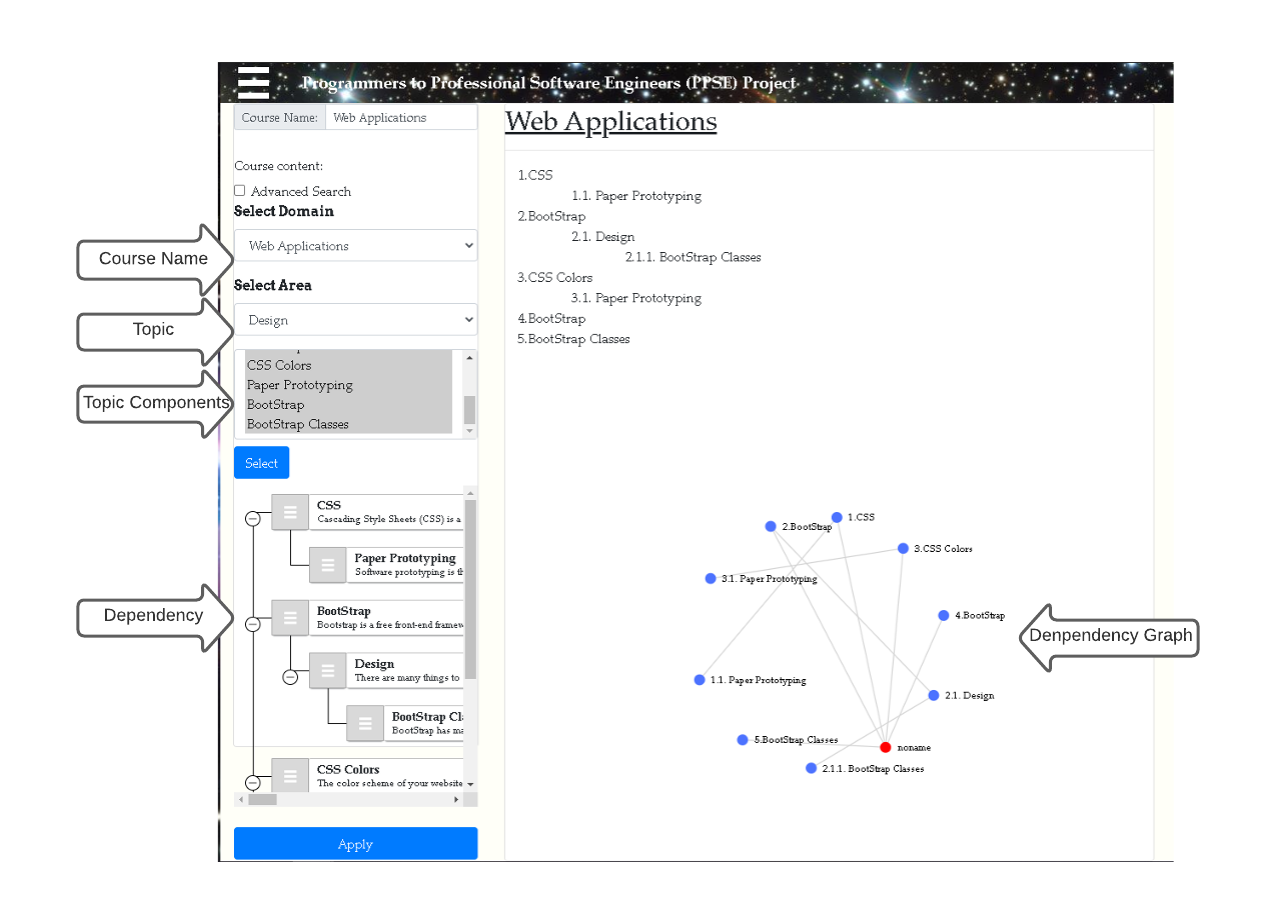}
\caption{Dependency Graph Creation: an instructor can choose topics (i.e., select area, see left top corner), select topic components, and design how to connect topic components (see left bottom corner) with a book chapter and sub-chapter style to define the dependencies.\label{fig:personalized}}
\end{figure}

To appeal to a wider spectrum of student interests, it is essential to choose diverse and varied topics. This can include various facets of robotics, such as mobile robotics, robotics navigation, and the physics underlying robotics. By providing a variety of topics, students can investigate several aspects of robotics and obtain a deeper understanding of the field. Consider the mini course's logical progression and intended learning outcomes when organizing the selected topics (see Figure~\ref{fig:personalized}). Start with topics that provide a solid comprehension of fundamental concepts and progress gradually to more advanced and specialized subjects. This progression enables students to build a solid foundation of knowledge and skills applicable to real-world situations. The selection and arrangement of topics for the mini-course can ensure a balanced curriculum that caters to the interests of students, encourages effective learning, and provides a solid foundation 

We have selected five core robotics subjects that are essential for a basic understanding of sensing and navigation problems. These include:
\begin{itemize}
    \item {\bf Sensors:} acquaints students with a wide array of sensors employed in the field of robotics, including but not limited to proximity sensors, cameras, LIDAR, and IMUs. This promotes the development of perception systems, which in turn facilitate effective interaction between robots and their surroundings; 
    \item {\bf Navigation:} is instrumental in enabling robots to independently traverse and orient themselves within their environment, a critical capability for a wide range of applications. Effective navigation leverages a robot's sensors to safely navigate in its environment;
    \item {\bf Dead Reckoning:} allows a robot to estimate its position and movement without being dependent on external localization systems. This skill is particularly useful in situations where these types of systems are either not available or not dependable; 
    \item  {\bf Potential Fields:} is one of many methodologies for path planning and obstacle avoidance, fundamental competencies for robots functioning in complex and dynamic surroundings; and
    \item {\bf Odometry:} the calculations from wheel encoders and sensors to calculate accurate location information.

\end{itemize}
    
\noindent The above topics provide a foundational exposure to robotics suitable for novice students to the field.

\subsection{Course Module Design}

The mini course's materials have been thoughtfully created to be beginner-friendly, allowing students to understand the topics with ease. 
Each module consists of two to three different sections including the reading section, the practical section, and the assessment or evaluation section. 

The reading section incorporates the objectives of instruction of the subject matter, a comprehensive overview of the topic with detailed explanations, and visual aids such as diagrams, charts, and infographics to enhance understanding and simplify complex ideas. 
We establish student learning objectives for students to delineate the intended knowledge and skills to be acquired, as well as the ultimate aim of the topic upon completion. We want to clearly present the linkage between the students' theoretical comprehension and practical applications of robotics.
This section leverages reference sources for follow-up and includes supplementary materials. 


Some topics have mathematical formulas that are necessary to understand the subject from its underlying theory. 
Students are also given additional information from outside sources during the course to help them better understand mathematical equations. Calculations are required for the concepts of potential field, odometry, and dead reckoning since they require a thorough understanding of the fundamental physics ideas. Students are given mathematical problems that have been solved and are then given equivalent activities to complete independently. For instance, after obtaining information on a robot's initial position, people can be asked to estimate the distance the robot has traveled in a given amount of time. An example would be to consider a two-wheeled robot that advances for five seconds. The left wheel rotates at a speed of ten revolutions per second, while the right wheel rotates at eight. The wheels have a 5-centimeter radius. What is the position and orientation of the robot?

The curriculum incorporates programming exercises to consolidate the fundamental concepts of the course into practical application of that knowledge. Programming examples and problems have been incorporated into the topics. When presented with the positions of an object, obstacle, and goal, students are required to determine the optimal path to reach the goal while avoiding the obstacle. 

\subsection{Student Evaluation}
Overall, the curriculum of the abridged course efforts to achieve a satisfactory balance between theoretical comprehension, mathematical principles, practical application, challenges, and evaluations. At the conclusion of each course, quizzes are administered as a means of assessing students' comprehension and progress. Through the utilization of these assessments, educators are able to evaluate the level of understanding of their students and pinpoint any areas that may necessitate additional clarification or reinforcement. The all-encompassing methodology guarantees that learners not only gain a strong theoretical basis but also practical proficiency, critical thinking skills, and the ability to apply their knowledge in real-life situations.

\section{Evaluation}

We recruited 16 participants from an Atlantic university campus to participate in the mini-course and to take a survey on their experience. Of these participants, 11 identified as male, 3 identified as female, and 2 preferred not to say. All students were fourth year students or higher; 4 were first-generation university students. When asked about racial backgrounds, 10 students identified as White (66\%), 1 as Hispanic (7\%), 2 as Black/African American (13\%), and 2 preferred not to say (13\%). All students expected to get an 'A' or 'B' in the course.

\textbf{Course Satisfaction:} Half of the students took the course out of interest in robotics. Students generally evaluated the course positively, with 88-94\% agreeing or strongly agreeing with positive general characteristics of the course and 81-88\% agreeing or strongly agreeing with positive items related to the course materials. Students were also asked about the course's impact on their plans related to robotics and their feelings about being a roboticist. Table 3 shows that 44\% of students were somewhat or extremely likely to go into robotics before taking the course, with 50\% reporting the same likelihood after the course. Additionally, 69\% of students agreed or strongly agreed that the course made them feel like a real roboticist. Students were also asked to provide open-ended feedback on the course, with many giving positive responses but noting glitches and revisions needed to the personalized learning system.

\textbf{Student Motivation:} The survey also included 12 items based on Self Determination Theory (SDT) to assess the extent to which the course supported students' autonomy, personal competence, and sense of relatedness to the class. In terms of autonomy, 63-100\% of students agreed or strongly agreed with items indicating that the course allowed them to make decisions about their learning. Between 75-93\% of students agreed or strongly agreed with items related to their ability to master course content, indicating a sense of competence. Between 80-93\% of students felt connected to the instructor, other students, and the class as a whole, indicating a sense of relatedness.

\section{Conclusions and Future Work}

We present an online learning system for self-selected learning for eventual deployment in community colleges, primarily undergraduate institutions, or other higher-education institutions where there is no robotics faculty member. The course model will hopefully facilitate student motivation and knowledge gain.

These preliminary results presented in this paper indicate that the course content presentation fosters both a sense of mastery of robotics content as well as engaging key motivational components of autonomy, competence, and relatedness. This is encouraging as one outcome of online courses can be a decrease in motivation to participate in course activities~\cite{syauqi2020students}. These results also show that the students were interested in the course content and enjoyed their participation in the mini-course.

Future work will resolve the technical issues identified above before the next round of student evaluations. Future evaluation work will also add a comparison of knowledge gained in robotics between online and in-person versions of the course to study whether this course model is effective for students in real classroom environments. While the size of the mini-course is likely too small to assess the effect of self-selection of topics for course content, future work will examine this question.
 
\bibliographystyle{ieeetr}
\bibliography{references}

\end{document}